\def\year{2019}\relax
\documentclass[letterpaper]{article} 
\usepackage{aaai19}  
\usepackage{times}  
\usepackage{helvet}  
\usepackage{courier}  
\usepackage{url}  
\usepackage{graphicx, array, blindtext}  
\usepackage{nicefrac} 
\usepackage{amsmath,amssymb,amsthm}
\usepackage{amsfonts}
\usepackage{algorithm}
\usepackage{subfigure}
\usepackage{algorithmic}

\usepackage{physics}
\usepackage{multirow}
\usepackage{longtable}
\usepackage{booktabs}
\usepackage{wrapfig}
\usepackage{caption}
\usepackage{float}
\usepackage{multirow}

\usepackage[caption = false]{subfig}

\begin{document}
%
\newtheorem{thm}{Theorem}
\newtheorem{lemma}{Lemma}
\newtheorem{corollary}{Corollary}
\newtheorem{remark}{Remark}
\newtheorem{proposition}{Proposition}
\newtheorem{example}{Example}
\newtheorem{definition}{Definition}

\title{FRAME Revisited: An Interpretation View Based on Particle Evolution}
\author{Xu Cai$^{1\dagger}$, Yang Wu$^{1\dagger}$, Guanbin Li$^1$, Ziliang Chen$^{1}$, Liang Lin$^{1,2}$\thanks{Xu Cai and Yang Wu contribute equally to this work and share first-authorship. Corresponding author is Liang Lin~(Email: linliang@ieee.org). This work was supported in part by the National Key Research and Development Program of China under Grant No.2018YFC0830103, in part by the NSFC-Shenzhen Robotics Projects~(U1613211), in part by the National Natural Science Foundation of China under Grant No.61702565, No.61622214 and No.61836012 and in part by National High Level Talents Special Support Plan (Ten Thousand Talents Program).} \\
$^1$School of Data and Computer Science, Sun Yat-Sen University, China\\
$^2$Dark Matter AI Inc.\\
\tt\small{caitree@foxmail.com,\ wuyang36@mail2.sysu.edu.cn,} \\
\tt\small{liguanbin@mail.sysu.edu.cn,\ c.ziliang@yahoo.com,\ linliang@ieee.org}
}
\maketitle

\begin{abstract}
FRAME~(Filters, Random fields, And Maximum Entropy) is an energy-based descriptive model that synthesizes visual realism by capturing mutual patterns from structural input signals. The maximum likelihood estimation~(MLE) is applied by default, yet conventionally causes the unstable training energy that wrecks the generated structures, which remains unexplained. In this paper, we provide a new theoretical insight to analyze FRAME, from a perspective of particle physics ascribing the weird phenomenon to KL-vanishing issue. In order to stabilize the energy dissipation, we propose an alternative Wasserstein distance in discrete time based on the conclusion that the Jordan-Kinderlehrer-Otto~(JKO) discrete flow approximates KL discrete flow when the time step size tends to $0$. Besides, this metric can still maintain the model's statistical consistency. Quantitative and qualitative experiments have been respectively conducted on several widely used datasets. The empirical studies have evidenced the effectiveness and superiority of our method.
 
\end{abstract}

\section{Introduction}
FRAME~(Filters, Random fields, And Maximum Entropy)~\cite{zhu1997minimax} is a model built on Markov random field that can be applied to approximate various types of data distributions, such as images, videos, audios and 3D shapes~\cite{lu2015learning,xie2017synthesizing,xie2018learning}. It is an energy-based descriptive model in the sense that besides its parameters are estimated, samples can be synthesized from the probability distribution the model specifies. Such distribution is derived from maximum entropy principle~(MEP), which is consistent with the statistical properties of the observed filter responses. FRAME can be trained via an information theoretical divergence between real data distribution $\mathbb{P}_{r}$ and model distribution $P_{\theta}$. Primitive efforts model it as KL-divergence by default, which also leads to the same results of MLE.

\begin{figure}[t]
    \centering
    \centerline{\includegraphics[width=\linewidth]{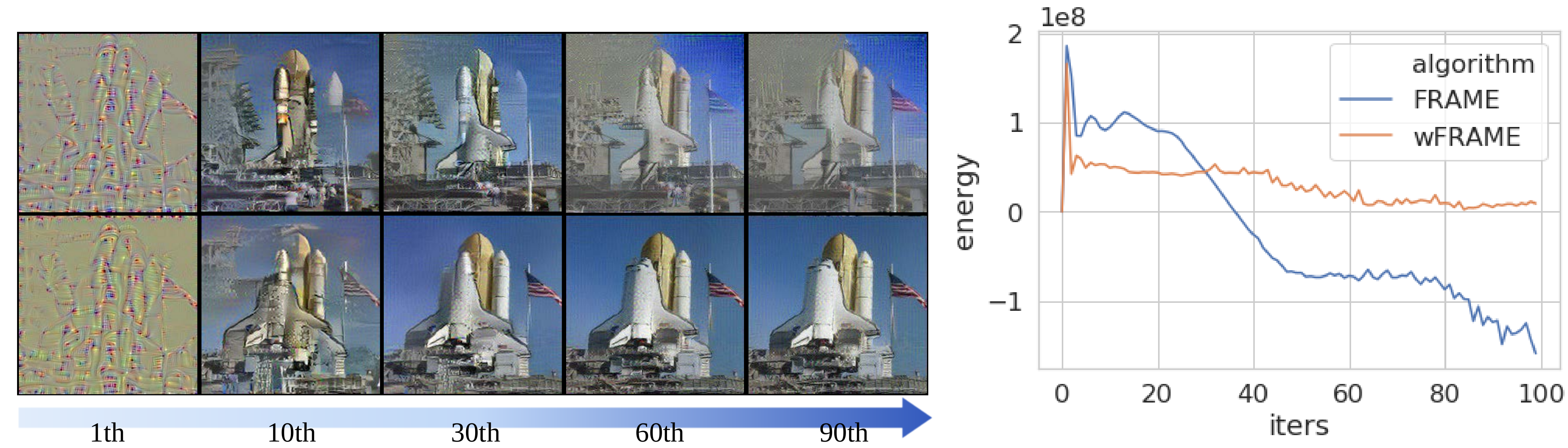}}
    \caption{Visual and numerical results of FRAME and wFRAME. Left: the generating steps and selected typical results of ``spaceship'' from two algorithms. The first and the second-row images are respectively from FRAME and wFRAME. wFRAME achieves higher quality images compared with FRAME, which collapses at the very beginning of the sampling iteration. Right: the observed model energy of both algorithms. The instability of the energy curve is the signal of the model collapse. The detailed discussion can be found in the experiment section.\label{fig:iters}}
\end{figure}

A large number of experimental results reveal that FRAME tends to generate inferior synthesized images and is often arduous to converge during training. For instance, displayed in Fig.~\ref{fig:iters}, the synthesized images of FRAME seriously deteriorates along with the model energy. This phenomenon is caused by KL-vanishing in the stepwise parameters estimation of the model due to the existence of the great filter responses disparity between $P_{\theta}$ and $\mathbb{P}_{r}$. Specifically, the MLE-based learning algorithm attempts to optimize a transformation from the high dimensional support of $P_{\theta}$ to the non-existing support of $\mathbb{P}_{r}$, i.e., it starts from an initialization of a Gaussian noise covering the whole support of $P_{\theta}$ and $\mathbb{P}_{r}$, then gradually updates $\theta$ by calculating the KL discrete flow step-wisely. Therefore in the discrete time setting of the actual iterative training process, the dissipation of the model energy may become considerably unstable, and the stepwise minimization scheme may suffer serious KL-vanishing issue during the communicative parameters estimation.

To tackle the above shortcomings, we first investigate this model from a particle perspective by regarding all the observed signals as Brownian particles~(pre-condition of KL discrete flow), which helps explore the reasons for the collapses of the FRAME model. This is inspired by the fact that the empirical measure of a set of Brownian particles generated by $P_{\theta}$ satisfies Large Deviation Principle~(LDP) with rate functional coincides exactly with the KL discrete flow~(see Lemma~\ref{lemma:1}). We then delve into the model in discrete time state and translate its learning mechanism from KL discrete flow into the Jordan-Kinderlehrer-Otto~(JKO)~\cite{jordan1998variational} discrete flow, which is a procedure for finding time-discrete approximations to solutions of diffusion equations in Wasserstein space. By resorting to the geometric distance between $P_{\theta}$ and $\mathbb{P}_{r}$ through optimal transport~(OT)~\cite{villani2003topics} and replacing the KL-divergence with Wasserstein distance~(a.k.a. the earth mover's distance~\cite{rubner2000earth}), this method manages to stabilize the energy dissipation scheme in FRAME and maintain its statistical consistency. The whole theoretical contribution can be summed up as the following deduction process:
\begin{itemize}
\item We deduce the learning process of data density in FRAME model from a view of particle evolution and confirm that it can be approximated by a discrete flow model with gradually decreasing energy driven by the minimization of the KL divergence.
\item We further propose Wasserstein perspective of FRAME~(wFRAME) by reformulating the FRAME's learning mechanism from KL discrete flow into the JKO discrete flow, of which the former theoretically explains the cause of the vanishing problem, while the latter overcomes the drawbacks, including the instability of sample generation and the failure of model convergence during training.
\end{itemize}

Qualitative and quantitative experiments demonstrate that the proposed wFRAME greatly ameliorates the vanishing issue of FRAME and can generate more visually promising results, especially for structurally complex training data. Moreover, to our knowledge, this method can be applied to most sampling processes which aim at abridging the KL-divergence between real data distribution and the generated data distribution by time sequence.

\section{Related Work}
\subsubsection{Descriptive Model for Generation.}
The descriptive models originated from statistical physics have an explicit probability distribution of the signal, where they are ordinarily called the Gibbs distributions~\cite{landau2013course}. With the massive developments of Convolutional Neural Networks~(CNN)~\cite{krizhevsky2012imagenet} which has been proven to be a powerful discriminator, recently, increasing researches on the generative perspective of this model have drawn a lot of attention.~\cite{dai2014generative} first introduces a generative gradient for pre-training discriminative ConvNet by a non-parametric importance sampling scheme and~\cite{lu2015learning} proposes to learn FRAME using pre-learned filters of modern CNN.~\cite{xie2016theory} further studies the theory of generative ConvNet intensively and show that the model has a representational structure which can be viewed as a hierarchical version of the FRAME model.
\subsubsection{Implicit Model for Generation.}
Apart from the descriptive models, another popular branch of deep generative models is black-box models which map the latent variables to signals via a top-down CNN, such as the Generative Adversarial Network~(GAN)~\cite{goodfellow2014} and its variants. These models have gained remarkable success in generating realistic images and learn the generator network with an assistant discriminator network.
\subsubsection{Relationship.}
Unlike the majority of implicit generative models, which use an auxiliary network to guide the training of the generator, descriptive models maintain a single model which simultaneously serves as a descriptor and generator, though FRAME can be served as an auxiliary and be combined with GAN to facilitate each other~\cite{xie2016cooperative}. They factually generate samples directly from the input set, rather than from the latent space, which to a certain extent ensures that the model can be efficiently trained and produce stable synthesized results with relatively less model structure complexity. In this paper, FRAME and its variants as described above share the same MLE based learning mechanism, which follows an analysis-by-synthesis scheme and works by first generating synthesized samples from the current model using Langevin dynamics and then learn the parameters through observed-synthesized samples' distance.

\section{Preliminaries}
\label{sec:pre}
Let $\mathcal{P}$ denote the space of Borel probability measures on any given subset of space $\mathcal{X}$, where $\forall\boldsymbol{x} \in \mathcal{X}$, $ \boldsymbol{x} \in \mathbb{R}^{d}$.
Given some sufficient statistics $\phi: \mathcal{X}\to\mathbb{R}$, scalar $\alpha\in\mathbb{R}$ and base measure $q$, the space of distributions satisfying linear constraint is defined as $\mathcal{P}_{\alpha}^{lin}=\left\{p,f\in\mathcal{P}: p=fq, f\geq 0, \int pdx=1, E_{p}[\phi(x)]=\alpha\right\}$. The Wasserstein space of order $r\in[1,\infty]$ is defined as $\mathcal{P}_{r}=\left\{p\in\mathcal{P}:\int\vert x\vert^{r}dp<\infty\right\}$, where $\vert\cdot\vert^{r}$ denotes the  $r$-norm on $\mathcal{X}$. $\vert \mathcal{X}\vert$ is the number of elements in domain $\mathcal{X}$. $\nabla$ denotes gradient and $\div$ denotes the divergence operator.

\subsubsection{Markov Random Fields~(MRF).}\label{sec:mrf} 
MRF belongs to the family of undirected graphical models, which can be written in the Gibbs form as
\begin{equation}
    \label{eq:mrf}
    P\left(\boldsymbol{x};\boldsymbol{\theta}\right) = \frac{1}{Z(\boldsymbol{\theta})}
    \exp\Bigg\{\sum_{k=1}^{K}\theta_{k}f_{k}(\boldsymbol{x})\Bigg\},
\end{equation}
where $K$ stands for the number of features $\left\{f_{k}\right\}_{k=1}^{K}$ and $Z(\cdot)$ is the partition function~\cite{koller2009probabilistic}. Its MLE learning process follows the iteration of the following two steps:

\textit{I.} Update model parameter $\boldsymbol{\theta}$ by ascending the gradient of the log likelihood
\begin{equation}
    \label{eq:mle}
    \frac{\partial}{\partial_{\theta_{k}}}\frac{1}{N}\log P\left(\boldsymbol{x};\boldsymbol{\theta}\right) = \mathbb{E}_{\mathbb{P}_{r}}[f_{k}\left(\boldsymbol{x}\right)] - \mathbb{E}_{P\left(\boldsymbol{x};\boldsymbol{\theta}\right)}\left[f_{k}    \left(\boldsymbol{x}\right)\right],
\end{equation}
where $\mathbb{E}_{\mathbb{P}_{r}}[f_{k}\left(\boldsymbol{x}\right)]$  and $\mathbb{E}_{P\left(\boldsymbol{x};\boldsymbol{\theta}\right)}\left[f_{k}    \left(\boldsymbol{x}\right)\right]$ is respectively the feature response over real data distribution $\mathbb{P}_{r}$ and current model distribution $P\left(\boldsymbol{x};\boldsymbol{\theta}\right)$.

\textit{II.} Sample from the current model by parallel MCMC chains. The sampling process, according to ~\cite{younes1989parametric}, does not necessarily converge at each $\boldsymbol{\theta}_{t}$, thus we only establish one persistent sampler that converges globally in order to reduce calculus.

\subsubsection{FRAME Model.} Based on an energy function, FRAME is defined on the exponential tilting of a reference distribution $q$, which is a reformulation of MRF and can be written as~\cite{lu2015learning}:
\begin{equation}
    \label{eq:standardframe}
    \begin{split}
        P\left(\boldsymbol{x};\boldsymbol{\theta}\right) & =
        \frac{1}{Z(\boldsymbol{\theta})}\exp\Bigg\{\sum_{k=1}^{K}\sum_{x\in\mathcal{X}}\theta_{k}h\left(\langle \boldsymbol{x}, \boldsymbol{w}\rangle+\boldsymbol{b}\right)_{k}\Bigg\}q(\boldsymbol{x}),
    \end{split}
\end{equation}
where $h(\boldsymbol{x})=\max(0,\boldsymbol{x})$ is the nonlinear activation function, $\langle\boldsymbol{x},\boldsymbol{w}\rangle$ is the filtered image or feature map and $q\left(\boldsymbol{x}\right)=\frac{1}{(2\pi\sigma^2)^{\vert \mathcal{X}\vert/2}} \exp\left[-\frac{1}{2\sigma^2}\Vert \boldsymbol{x}\Vert^2\right]$ denotes the Gaussian white noise model with mean $0$ and variance $\sigma^{2}$.

\subsubsection{KL Discrete Flow.} This flow is related to discrete probability distributions~(evolutions discretized in time) with finite dimensional problems. More precisely,  it indicates the system of $n$ independent Brownian particles $\{\boldsymbol{x}^{i}\}_{i=1}^{n} \in\mathbb{R}^{d}$ whose position in $\mathbb{R}^{d}$ is given by a Wiener process satisfies the following stochastic differential equation~(SDE)
\begin{equation}
    \label{eq:ito}
    d\boldsymbol{x}_{t} = \mu(\boldsymbol{x}_{t})dt + \varepsilon(\boldsymbol{x}_{t})d\boldsymbol{B}_{t}.
\end{equation}
$\mu$ is the drift term, $\varepsilon$ stands for the diffusion term, $\boldsymbol{B}$ denotes the Wiener process and subscript $t$ denotes time point $t$.
This empirical measure of those particles is proved to approximate Eq.~\ref{eq:standardframe} by an implicit descent step $\rho^{*}=\operatorname*{argmin}_{\rho}\mathcal{I}_{t}$, where $\mathcal{I}_{t}$ is the so called KL discrete flow consists of KL divergence and energy function $\Phi:\mathbb{R}^{d}\to\mathbb{R}$.
\begin{equation}
    \label{eq:klflow}
    \mathcal{I}_{t} = \mathcal{K}(\rho\mid\rho_{t}) + \int \Phi d\rho.
\end{equation}

\section{Particle Perspective of FRAME Model}
Although there is a traditional statistical perspective to interpret the FRAME theory~\cite{xie2016theory}, we still need a more stable sampling process to avoid this frequent generation failure. We revisit the frame model from a completely new particle perspective and prove that its parameter update mechanism is actually equivalent to the reformulation of KL discrete flow. Its further transformation, a mechanism in JKO discrete flow manner which we will next prove the equivalence on condition of enough sampling time steps, has ameliorated this unpredictably vanishing phenomenon. All the proofs in detail are added to Appendix A.

\subsection{Discrete Flow Driven by KL-divergence}
\label{sec:klflow}
Herein we first introduce FRAME in discrete flow manner. If we regard the observed signals $\{\boldsymbol{x}^{i}_{t}\}_{i=1}^{n}$ with the generating function of Markov property as Brownian particles, then theorem~\ref{thm:klflow} points out that Langevin dynamics can be deduced from KL discrete flow sufficiently and necessarily through lemma~\ref{lemma:1}. 
\begin{lemma}
    \label{lemma:1}
    For i.i.d. particles $\{\boldsymbol{x}^{i}_{t}\}_{i=1}^{n}$ with common generating function $\mathbb{E}[e^{\Phi(\boldsymbol{x};\boldsymbol{\theta})}]$ which has Markov property, the empirical measure $\rho_{t}=\frac{1}{n}\sum_{i=1}^{n}\delta_{\boldsymbol{x}^{i}_{t}}$ satisfies Large Deviation Principle~(LDP) with rate functional in the form of $\mathcal{I}_{t}$.
\end{lemma}

\begin{thm}
    \label{thm:klflow}
    Given a base measure $q$, a clique potential $\Phi$ , the density of FRAME in Eq.~\ref{eq:standardframe} can be obtained \textbf{sufficiently} and \textbf{necessarily} by solving the following constrained optimization.
    \begin{equation}
        \begin{split}
            \label{eq:klconstrain}
            \rho_{t+1} &= \operatorname*{argmin}_{\rho}\mathcal{K}(\rho\mid\rho_{t}), \\
            s.t. \int\Phi d\rho =\int\Phi &d\mathbb{P}_{r}, \quad \rho_{0}=q, \quad\forall\rho\in\mathcal{P}_{\alpha}^{lin}.
        \end{split}
    \end{equation}
\end{thm}
Let $\boldsymbol{\theta}$ be the Lagrange multiplier integrated in $\Phi(\boldsymbol{x};\boldsymbol{\theta})$ and ensure $\mathbb{E}[e^{\Phi(\boldsymbol{x};\boldsymbol{\theta})}]<\infty$, then the optimizing objective can be reformulated as 
\begin{equation}
    \begin{split}
        \label{eq:klflow_1}
        & \mathcal{I}_{t}^{l} =\operatorname*{min}_{\rho}\operatorname*{max}_{\boldsymbol{\theta
        }} \Big\{\mathcal{K}(\rho\mid\rho_{t}) + \int\Phi(\boldsymbol{x};\boldsymbol{\theta
        }) d\rho - \int\Phi(\boldsymbol{x};\boldsymbol{\theta
        }) d\mathbb{P}_{r}\Big\}.
    \end{split}
\end{equation}

Since $\nabla_{\boldsymbol{x}}\log P(\boldsymbol{x};\boldsymbol{\theta})=\nabla_{\boldsymbol{x}}\Phi(\boldsymbol{x};\boldsymbol{\theta})$, then the SDE iteration of $\boldsymbol{x}_{t}$ in Eq.~\ref{eq:ito} can be expressed in the Langevin form as
\begin{equation}
    \label{eq:langevin}
    \boldsymbol{x}_{t+1} = \boldsymbol{x}_{t} + \nabla_{\boldsymbol{x}}\log P(\boldsymbol{x}_{t};\boldsymbol{\theta}) + \sqrt{2}\boldsymbol{\xi}_{t}.
\end{equation}
By Lemma~\ref{lemma:1}, if we fix $\boldsymbol{\theta}$, the sampling scheme in Eq.~\ref{eq:langevin} approaches the KL discrete flow $\mathcal{I}_{t}^{l}$,
the flow will fluctuate in case $\boldsymbol{\theta}$ varies. $\boldsymbol{\theta}$ is updated by calculating $\nabla_{\theta}\mathcal{I}_{t}^{l}$, which implies $\boldsymbol{\theta}$ can dynamically transform the transition map into desired. The sampling process of FRAME can be summed up as
\begin{equation}
    \label{eq:standard_learning}
    \left\{
    \begin{aligned}
        \boldsymbol{x}_{t+1}      & = \boldsymbol{x}_{t} - \left(\frac{\boldsymbol{x}_{t}}{\sigma^{2}} - \nabla_{\boldsymbol{x}}{\Phi(\boldsymbol{x}_{t};\boldsymbol{\theta})}\right) + \sqrt{2}\boldsymbol{\xi}_{t} \\
        \boldsymbol{\theta}_{t+1} & = \boldsymbol{\theta}_{t} + \nabla_{\boldsymbol{\theta}}\mathbb{E}_{\rho_{t}}[\Phi(\boldsymbol{x};\boldsymbol{\theta})] -\nabla_{\boldsymbol{\theta
            }}\mathbb{E}_{\mathbb{P}_{r}}[\Phi(\boldsymbol{x};\boldsymbol{\theta})],
    \end{aligned}
    \right.
\end{equation}
where $-\boldsymbol{x}_{t}/\sigma^{2}$ is the derivative of initial Gaussian noise $q$. If we take a close look at the objective function, there is an
adversarial mechanism while updating $\boldsymbol{x}_{t}$ and $\boldsymbol{\theta}_{t}$. Regardless of fixing $\boldsymbol{\theta}$ updating $\boldsymbol{x}$, or fixing $\boldsymbol{x}$ updating $\boldsymbol{\theta}$, the correct direction cannot be insured to the optimal of minimizing $\mathcal{K}(P(\boldsymbol{x};\theta)\mid\mathbb{P}_{r})$.

\subsection{Discrete Flow Driven by Wasserstein Metric}
\label{sec:jkoflow}
Although KL approach is relatively rational in the methodology of FRAME, there exists the risk of a KL-vanishing problem as we have discussed, since the parameter updating mechanism of MLE may suffer non-convergence. To avoid this problem, we introduce the Wasserstein metric to discrete flow, according to the statement of~\cite{montavon2016wasserstein} that $P_{\theta}$ can be closer from a KL method given empirical measure $\rho_{t}$, but far from the same measure in the Wasserstein distance. And~\cite{arjovsky2017wasserstein} also claims that a better convergence and approximated results can be obtained since Wasserstein metric defines a weaker topology. The conclusion that $\mathcal{I}_{t} \approx \mathcal{J}_{t}$ when time step size $\tau \to 0$ rationalizes the proposed method. The proof of this conclusion in the one-dimensional situation has shown in~\cite{adams2011large} and in higher-dimensional has been proved by~\cite{duong2013wasserstein,erbar2015large}. Here we first provide some background knowledge about the transformation then we briefly show the derivation process.

\subsubsection{Fokker-Planck Equation.} Under the influence of drifts and random diffusions, this equation describes the evolution for the probability density function of the particle velocity. Let $F$ be an integral function and $\delta F/\delta \rho$ denote its Euler-Lagrange first variation, the equations are
\begin{equation}
    \label{eq:fokker}
    \left\{
    \begin{aligned}
        &\partial_{t} \rho + \div(\rho \nu) =0 \quad                                  & \text{(Continuity equation)}                     \\
        &\nu=-\nabla\frac{\delta F}{\delta\rho} \quad & \text{(Variational condition)}                   \\
        & \rho(\cdot,0)=\rho_{0} \quad                           & \rho_{0} \in L^{1}(\mathbb{R}^d),\rho_{0}\geq 0.
    \end{aligned}
    \right.
\end{equation}

\subsubsection{Wasserstein Metric.}
The Benamou-Brenier form of this metric~\cite{benamou2000computational} of order $r$ involves solving a smoothy OT problem over any probabilities $\mu_{1}$ and $\mu_{2}$ in $\mathcal{P}_{r}$ using the continuity equation showed in Eq.~\ref{eq:fokker} as follows, where $\nu$ belongs to the tangent space of the manifold governed by some potential and associated with curve $\rho_{t}$. 
\begin{equation}
    \label{eq:w_metric_2}
    \begin{split}
        \mathcal{W}^{r}(\mu_{1}, \mu_{2}) := \operatorname*{min}_{\rho_{t}\in\mathcal{P}_{r}}
  \{  \int_{0}^{1} \int_{\mathbb{R}^{d}}|\nu_{t}|^{r}d\rho_{t} dt : \partial_{t}\rho_{t} \\
  + \div(\rho_{t} \cdot \nu_{t})=0~|~\rho_{0} = \mu_{1}, \rho_{1}=\mu_{2}\}.
    \end{split}
\end{equation}

\subsubsection{JKO Discrete Flow.}
Following the initial work~\cite{jordan1998variational}, which shows how to recover Fokker-Planck diffusions of distributions in Eq.~\ref{eq:fokker} when minimizing entropy functionals according to Wasserstein metric $\mathcal{W}^{2}$, the JKO discrete flow is applied by our method to replace the initial KL divergence with the entropic Wasserstein distance $\mathcal{W}^{2}-H(\rho)$. The function of the flow is 
\begin{equation}
    \label{eq:jkoflow}
    \mathcal{J}_{t} = \frac{1}{2} \mathcal{W}^{2}(\rho,\rho_{t}) +
    \int \log\rho d\rho  + \int\Phi d\rho.
\end{equation}

\begin{remark}
    \label{remark:jkoflow}
    The initial Gaussian term $q$ is left out for convenience to facilitate the derivation, otherwise, the entropy $-H(\rho)=\int \log\rho d\rho$ in Eq.~\ref{eq:jkoflow} should be written as the relative entropy $\mathcal{K}(\rho\mid q)$.
    \end{remark}

By Theorem~\ref{thm:klflow}, $\mathcal{J}_{t}$ instead of $\mathcal{I}_{t}$ can be calculated in approximation and a steady state will approach Eq.~\ref{eq:standardframe}. Applying $\mathcal{J}_{t}$ in the manner of dissipation mechanism as a substitute of $\mathcal{I}_{t}$ allows regarding the diffusion Eq.~\ref{eq:ito} as the steepest descent of clique energy $\Phi$ and entropy $-H(P)$ w.r.t. Wasserstein metric. Solving such optimization problem using $\mathcal{W}$ is identical to solve the Monge-Kantorovich mass transference problem. 

With Second Mean Value theorem for definite integrals, we can approximately recover the integral $\mathcal{W}^{2}$ by two randomly interpolated rectangles
\begin{equation}
\label{eq:interpolate}
\begin{split}
    &\mathcal{W}^{2}\left(\rho_{t_{0}}, \rho_{t_{1}}\right) :=
        \operatorname*{inf}_{\rho_{t}}\int_{t_{0}}^{t_{1}}
    \int_{\mathbb{R}^{d}}\vert\nabla\Phi\vert^{2}d\rho_{t} dt \\
    &\approx 
  \left(\zeta-t_{0}\right)\int_{\mathbb{R}^{d}}\vert\nabla\Phi\vert^{2}d\rho_{t_{0}} +
    \left(t_{1}-\zeta\right)\int_{\mathbb{R}^{d}}\vert\nabla\Phi\vert^{2}d\rho_{t_{1}} \\
        & = 
        -\beta\left(\left(1-\gamma\right)\int_{\mathbb{R}^{d}}\vert\nabla\Phi\vert^{2}d\rho_{t_{0}} + \gamma\int_{\mathbb{R}^{d}}\vert\nabla\Phi\vert^{2}d\rho_{t_{1}}\right).
    \end{split}
\end{equation}
where $\beta=t_{1}-t_{0}$ parameterizes the time piece and $\gamma=\zeta/\beta~(0\leq\gamma\leq1)$ represents random interpolated parameter since $\zeta$ is random. With Eq.~\ref{eq:interpolate}, the functional derivative of $\mathcal{W}^{2}(\rho_{t_{0}}, \rho_{t_{1}})$ w.r.t. $\rho_{t_{1}}$ is then proportional to
\begin{equation}
    \label{eq:derive_wmetric}
    \frac{\delta \mathcal{W}^{2}(\rho_{t_{0}}, \rho_{t_{1}})}{\delta\rho_{t_{1}}} \propto \vert\nabla\Phi\vert^{2},
\end{equation}
which is exactly the result of Proposition 8.5.6 in~\cite{ambrosio2008gradient}. Assume $\Phi$ be at least twice differentiable and treat Eq.~\ref{eq:derive_wmetric} as the variational condition in Eq.~\ref{eq:fokker}, then plug Eq.~\ref{eq:derive_wmetric} into the continuity equation of Eq.~\ref{eq:fokker}, which turns into a modified Wasserstein gradient flow in Fokker-Planck form as follows
\begin{equation}
    \label{eq:new_fokker}
    \partial_{t} \rho = \Delta \rho -
    \div(\rho (\nabla \Phi - \nabla \vert \nabla \Phi(\boldsymbol{x}) \vert^{2})).
\end{equation}
Then the corresponding SDE can be written in Euler-Maruyama form as
\begin{equation}
    \label{eq:new_langevin}
    \boldsymbol{x}_{t+1} = \boldsymbol{x}_{t} + \nabla{\Phi(\boldsymbol{x}_{t})} - \nabla\vert\nabla\Phi(\boldsymbol{x}_{t})\vert^{2} + \sqrt{2}\boldsymbol{\xi}_{t}.
\end{equation}
By Remark~\ref{remark:jkoflow}, if we reconsider the initial Gaussian term, the discrete flow of $\boldsymbol{x}_{t+1}$ in Eq.~\ref{eq:new_langevin} should be added with $-\boldsymbol{x}_{t}/\sigma^{2}$.
\begin{remark}
    \label{remark:0}
    If $\Phi$ is the energy function defined in Eq.~\ref{eq:standardframe}, then $\nabla\vert\nabla\Phi(\boldsymbol{x})\vert^{2}=0$.
\end{remark}

It's a direct result since $\Phi(\boldsymbol{x},\boldsymbol{\theta})$ defined in FRAME  only involves inner-product, ReLu~(piecewise linear) and other linear operations, the second derivative is obviously $0$. Therefore, both the time evolution of density $\rho_{t}$ in Eq.~\ref{eq:new_fokker} and sample $\boldsymbol{x}_{t}$ in Eq.~\ref{eq:new_langevin} will respectively degenerate to Eq.~\ref{eq:fokker} and Eq.~\ref{eq:langevin}. Thus the SDE of $\boldsymbol{x}_{t}$ remains default, i.e. Langevin form while the gradients of the model parameter $\boldsymbol{\theta}_{t}$ doesn't degenerate.

Alike to the parameterized KL flow $\mathcal{I}_{t}^{l}$ defined in Eq.~\ref{eq:klflow_1}, we propose a similar form in JKO manner. With Eq.~\ref{eq:interpolate} and Eq.~\ref{eq:derive_wmetric}, the \textbf{final optimization objective function} $\mathcal{J}_{t}^{l}$ can be formulated as
\begin{equation}
    \begin{split}
        \mathcal{J}_{t}^{l} = & \operatorname*{min}_{\rho}\operatorname*{max}_{\boldsymbol{\theta}}
        \Big\{ -\frac{\beta}{2}\left(1-\gamma\right)\int_{\mathbb{R}^{d}}\vert\nabla_{\boldsymbol{x}}\Phi(\boldsymbol{x};\boldsymbol{\theta})\vert^{2}d\rho_{t} \\
        & - \frac{\beta}{2}\gamma\int_{\mathbb{R}^{d}}\vert\nabla_{\boldsymbol{x}}\Phi(\boldsymbol{x};\boldsymbol{\theta})\vert^{2}d\rho
    + \int log\rho d\rho \\
    & +\int\Phi(\boldsymbol{x};\boldsymbol{\theta}) d\rho -
        \int\Phi(\boldsymbol{x};\boldsymbol{\theta}) d\mathbb{P}_{r}\Big\}.
\end{split}
\end{equation}

With all discussed above, the learning progress of wFRAME can be constructed by ascending the gradient of $\theta$, i.e. $\nabla_{\boldsymbol{\theta}}\mathcal{J}_{t}^{l}$. The calculating steps in formulation are summarized in Eq.~\ref{eq:new_learning}.
\begin{equation}
    \label{eq:new_learning}
    \left\{
    \begin{aligned}
        \boldsymbol{x}_{t+1} & = \boldsymbol{x}_{t} - \left(\frac{\boldsymbol{x}_{t}}{\sigma^{2}} - \nabla_{\boldsymbol{x}}{\Phi(\boldsymbol{x}_{t};\boldsymbol{\theta})}\right) + \sqrt{2}\boldsymbol{\xi}_{t} \\
        \boldsymbol{\theta
        }_{t+1}              & = \boldsymbol{\theta}_{t} + \nabla_{\boldsymbol{\theta}}\mathbb{E}_{\rho_{t}}[\Phi(\boldsymbol{x};\boldsymbol{\theta})]
        -\nabla_{\boldsymbol{\theta}}\mathbb{E}_{\mathbb{P}_{r}}[\Phi(\boldsymbol{x};\boldsymbol{\theta})] \\
                                & - \frac{\beta}{2}\left(1-\gamma\right)\nabla_{\boldsymbol{\theta}}\mathbb{E}_{\rho_{t-1}}
        [\vert\nabla_{\boldsymbol{x}}\Phi(\boldsymbol{x};\boldsymbol{\theta})\vert^{2}] \\
                &    - \frac{\beta}{2}\gamma\nabla_{\boldsymbol{\theta}}\mathbb{E}_{\rho_{t}}
        [\vert\nabla_{\boldsymbol{x}}\Phi(\boldsymbol{x};\boldsymbol{\theta
        })\vert^{2}].
    \end{aligned}
    \right.
\end{equation}
The equation above indicates that the gradient of $\boldsymbol{\theta}$ in Wasserstein manner is being added with some soft gradient norm constraints between the last two iterations. Such gradient norm has the following \textbf{advantages} compared with the original iteration process~(Eq.~\ref{eq:standard_learning}).

First the norm serves as the constant speed geodesic connecting $\rho_{t}$ with $\rho_{t+1}$ in the manifold spanned by $P_{\boldsymbol{\theta}}$ and $\mathbb{P}_{r}$, which may provide a speedup on converge. Next, it can be interpreted as the soft anti-force against the original gradient and prevent the whole learning process from vanishing. Moreover, in experiments, we find it can preserve data inner structural information. The new learning and generating the process of wFRAME is summarized in Algorithm~\ref{alg:Wasserstein FRAME} in detail.
\setlength{\textfloatsep}{0.5 cm}
\begin{algorithm}[t!]
    \renewcommand{\algorithmicrequire}{\textbf{Input:}}
    \renewcommand{\algorithmicensure}{\textbf{Output:}}
    \caption{Persistent Learning and Generating in Wasserstein FRAME}
    \label{alg:Wasserstein FRAME}
    \begin{algorithmic}[1]    
        \REQUIRE  Training data $\{\boldsymbol{y}^i,i=1,...,N\}$
        \ENSURE   Synthesized data $\{\boldsymbol{x}^i,i=1,...,M\}$
        \STATE  Initialize $\boldsymbol{x}_{0}^i \leftarrow 0$
        \FOR{$t=1$ {\bfseries to} $T$}
        \STATE  $H^{obs}\leftarrow\frac{1}{N}\sum_{i}^{N}\nabla_{\boldsymbol{\theta}_{t}}\Phi(\boldsymbol{y}^i)$
        \FOR{$j=1$ {\bfseries to} $L$}
        \STATE $\mathcal{G}\leftarrow \nabla_{\boldsymbol{x}_{t\times L + j-1}}\Phi(\boldsymbol{x}_{t\times L + j-1})$
        \STATE $\mathcal{S}\leftarrow \frac{\boldsymbol{x}_{t\times L + j-1}}{\sigma^2}$
        \STATE Sample $\boldsymbol{\Sigma}\leftarrow \mathcal{N}(0,\sigma^{2}\cdot\mathbf{I}_{d})$
        \STATE $\boldsymbol{x}_{t\times L + j} \leftarrow \boldsymbol{x}_{t\times L + j-1} +
            \frac{\delta^2}{2}(\mathcal{G}-\mathcal{S}) + \delta \boldsymbol{\Sigma}$
        \ENDFOR
        \STATE  $H^{syn}\leftarrow\frac{1}{M}\sum_{i}^{M}\nabla_{\boldsymbol{\theta}_{t}}\Phi(\boldsymbol{x}_{(t+1)\times L}^i)$
        \STATE  $\mathcal{P}_{t}\leftarrow \frac{1}{M}\sum_{i}^{M}
            \nabla_{\boldsymbol{\theta}_{t}}\vert\nabla_{\boldsymbol{x}_{(t+1)\times L}}\Phi(\boldsymbol{x}_{(t+1)\times L}^i)\vert^2$
        \STATE   $\mathcal{P}_{t-1}\leftarrow \frac{1}{M}\sum_{i}^{M}
            \nabla_{\boldsymbol{\theta}_{t}}\vert\nabla_{\boldsymbol{x}_{t\times L }}\Phi(\boldsymbol{x}_{t\times L }^i)\vert^2$
        \STATE Sample $\gamma \sim U[0,1]$
        \STATE Update $\boldsymbol{\theta}_{t+1} \leftarrow \boldsymbol{\theta}_{t} +
            \lambda\cdot(H^{obs}-H^{syn}) - \frac{\beta}{2}\left(\left(1-\gamma\right)\mathcal{P}_{t-1} + \gamma\mathcal{P}_{t}\right)$
        \ENDFOR
    \end{algorithmic}
\end{algorithm}

\section{Experiments}
\label{sec:experiment}
In this section, we intensively compare our proposed method with FRAME from two aspects, one is the confirmatory experiment of model collapse under varied settings with respect to the baseline, the other is the quantitative and qualitative comparison of generated results on extensively used datasets. In the first stage, as expected, the proposed wFRAME is verified to be more robust in training and the synthesized images are of higher quality and fidelity in most circumstances. The second stage, we evaluate both models on the whole datasets. We propose a new metric response distance, which measures the gap between the generated data distribution and the real data distribution.

\subsection{Confirmation of Model Collapse}
We recognize that under some circumstances FRAME will suffer serious model collapse. Due to MEP, the expected well-learned FRAME model $P_{\boldsymbol{\theta}}^{*}$ should achieve minimum $\mathcal{K}(P_{\boldsymbol{\theta}}^{*}\mid q)$, i.e. the minimum amount of transformations to the reference measure. But such minimization of KL divergence might be the unpredictable cause of the energy to $0$, namely the learned model will degenerate to produce initial noise instead of the desired minimum modification. Furthermore, in case $\Phi(\boldsymbol{x},\boldsymbol{\theta}) \leq 0$, the learned model intends to degenerate. In other words, the images synthesized from FRAME driven by KL divergence will collapse immediately and the quality may barely restore. Consequently, the best curve of $\Phi$ is slowly asymptotic to and slightly above $0$.

\begin{figure*}[t!]
    \centering
    \includegraphics[width=\linewidth]{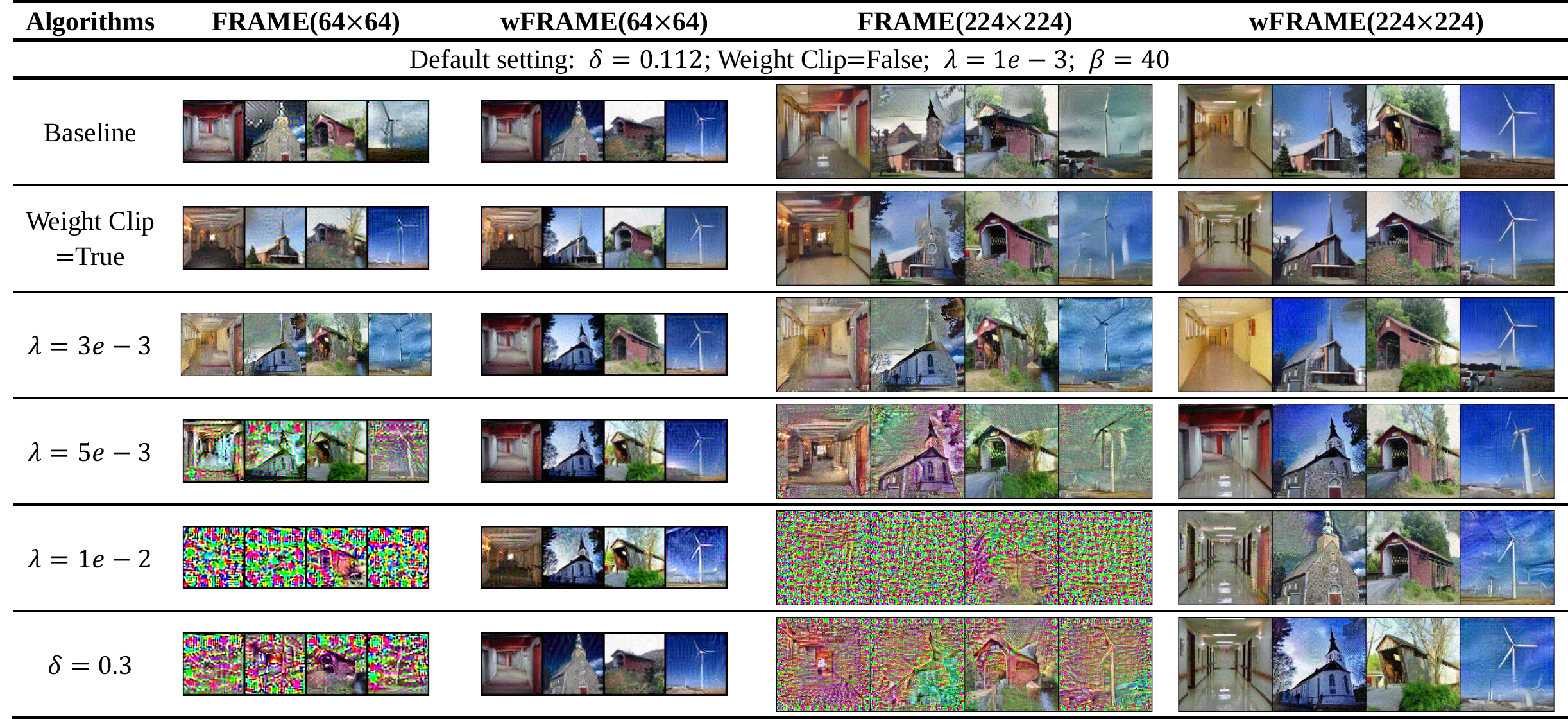}
    \caption{The synthesized results under different circumstances.\label{fig:samples}}
\end{figure*}

To manifest the superiority of our method over FRAME compared with the baseline settings, we conduct the validation experiments on a subset of SUN dataset~\cite{xiao2010sun} under different circumstances. Intuitively, a simple trick to the model collapse issue is to restrict $\boldsymbol{\theta}$ in a safe range, a.k.a. weight clipping. The experimental settings include respectively altering $\lambda$ and $\delta$ to an insecure range, turning on or off the weight clipping and varying the inputs dimensions. The results are presented in Fig.~\ref{fig:samples}, which shows the property of a more robust generation compared with the original strategy or FRAME with weight clipping trick.

\subsection{Empirical Setup on Common Datasets}
\begin{figure}[b!]
    \centering
    \includegraphics[width=\linewidth]{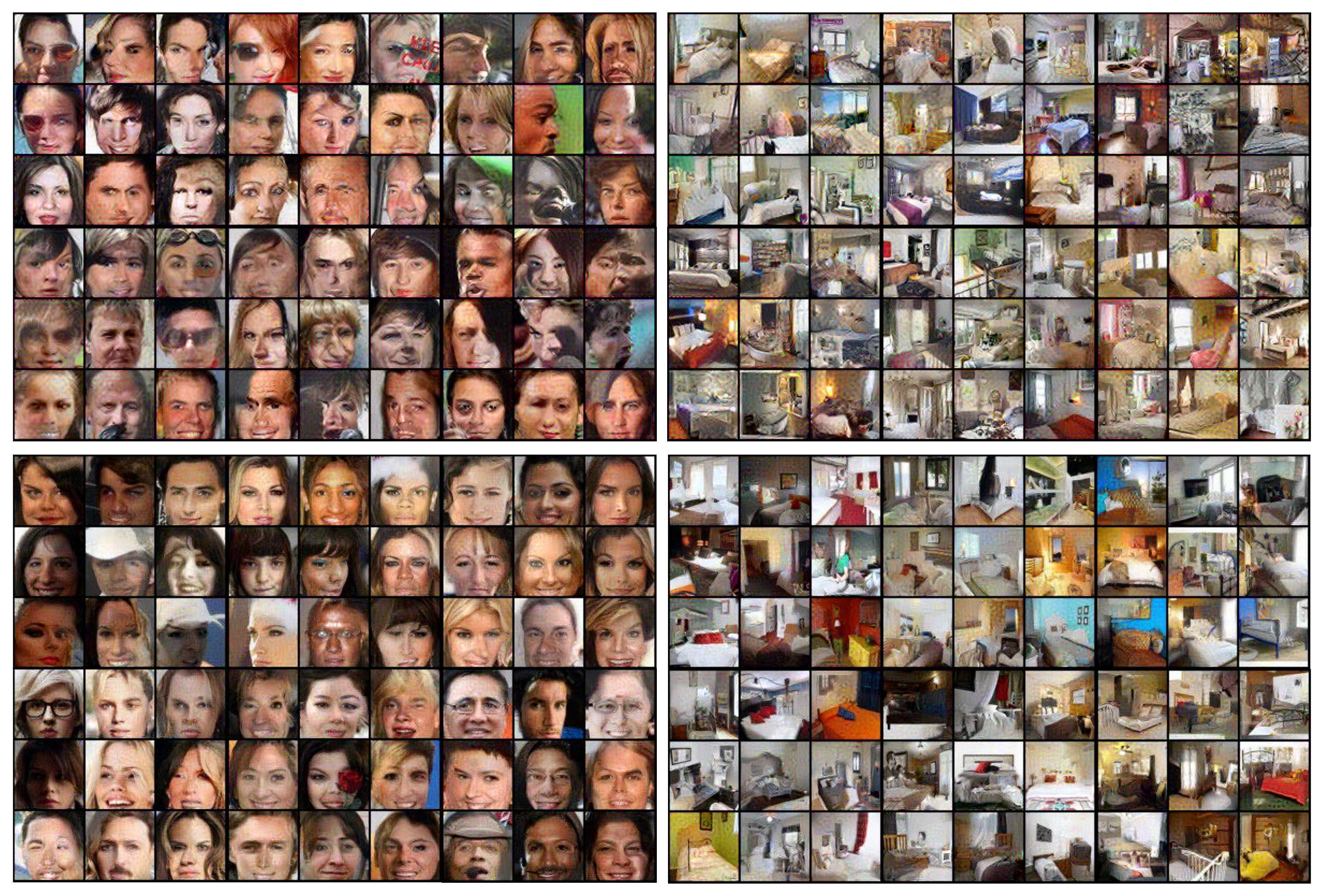}
    \caption{Comparison on LSUN-Bedroom and CelebA, where the first row is synthesized from FRAME the second is from wFRAME. More visual results have been added to the Appendix B.\label{fig:celeba_lsun}}
\end{figure}

We apply wFRAME on several widely used datasets in the field of generative modeling. As for default experimental settings, $\sigma=0.01$, $\beta=60$, the number of learning iterations is set to $T=100$, the step number $L$ of Langevin sampling within each learning iteration is $50$ and the batch size is $N=M=9$. The implementation of $\Phi(x)$ in our method is the first 4 convolutional layers of a pre-learned VGG-16~\cite{simonyan2014very}. Input shape varies by datasets and is specified following. The hyper-parameters appear in Algorithm~\ref{alg:Wasserstein FRAME} differs on each dataset in order to achieve the best results. As for FRAME we use default settings in~\cite{lu2015learning}.

\textbf{CelebA}~\cite{liu2015deep} and \textbf{LSUN-Bedroom}~\cite{yu15lsun} images are cropped and resized to $64\times64$. we set $\lambda=1e^{-3}$ in both datasets, $\delta=0.2$ in CelebA and $\delta=0.15$ in LSUN-Bedroom. The visualizations of two methods are exhibited in Fig.~\ref{fig:celeba_lsun}.

\begin{figure}[h!]
    \centering
    \centerline{\includegraphics[width=\linewidth]{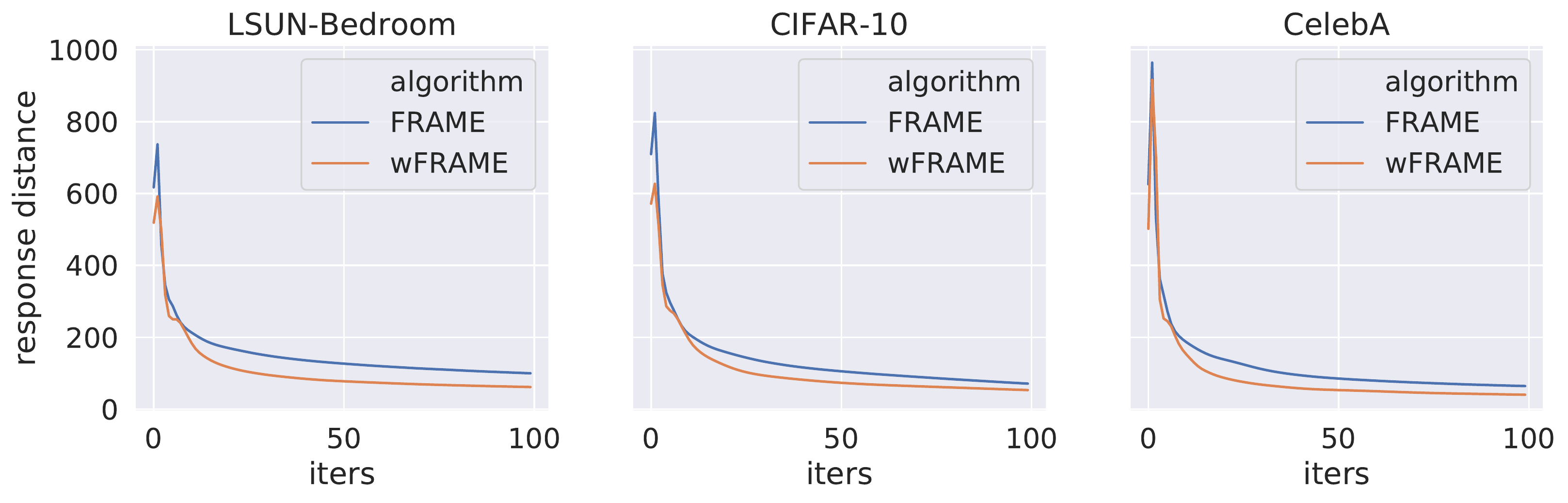}}
    \caption{The averaged learning curves of response distance $R$ on  CelebA, LSUN-Bedroom and CIFAR-10.\label{fig:curve}}
\end{figure}

\begin{figure*}
    \centering
    \centerline{\includegraphics[width=0.8\paperwidth]{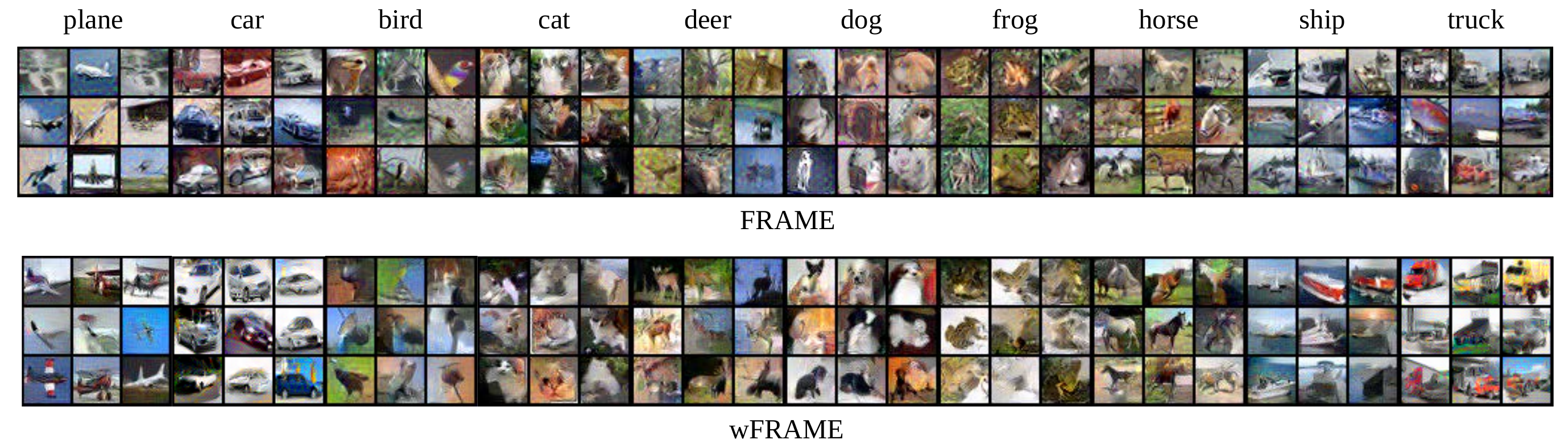}}
    \caption{Images generated by two algorithms conditioned on labels in CIFAR-10, every three columns are of one class, the first group is from FRAME and the second is from wFRAME.\label{fig:cifar_vis}}
\end{figure*}

\textbf{CIFAR-10}~\cite{krizhevsky2009learning} includes various categories and we learn both algorithms conditioned on the class label. In this experiment, we set $\delta=0.15$, $\lambda=2e^{-3}$ and images' size are of $32\times32$. Numerically and visually in Fig.~\ref{fig:curve},~\ref{fig:cifar_vis} and Table~\ref{tab:cifar}, the results show great improvement.

For a fair comparison, two metrics are utilized to evaluate FRAME and wFRAME. We offer a new metric response distance to measure the disparity between two distributions according to the results sampled out, while the Inception score is a widely used standard in measuring samples diversity.

\textbf{Response distance} $R$ is defined as

\begin{equation*}
    R = \frac{1}{K}\sum_{k=1}^{K}\left\lvert\frac{1}{N}\sum_{i=1}^{N}F_{k}(\boldsymbol{x}^{i}) - \frac{1}{M}\sum_{i=1}^{M}F_{k}(\boldsymbol{y}^{i})\right\rvert
\end{equation*}
where $F_{k}$ denotes the $k$th filter. The smaller the $R$ is, the better the generated results will be, since $R\propto{max}_{\theta}\mathbb{E}_{r}[F(\boldsymbol{y}^{i})]-\mathbb{E}_{P_{\theta}}[F(\boldsymbol{x}^{i})]$, which implies that $R$ provides an approximation of the divergence between the target data distribution and the generated data distribution. Furthermore, by Eq.~\ref{eq:mle}, the faster $R$ falls the better $\boldsymbol{\theta}$ converges.

\textbf{Inception score~(IS)} is the most widely adopted metric of generative models, which estimates the diversity of the generated samples. It uses a network Inception v2~\cite{szegedy2016rethinking} pre-trained on ImageNet~\cite{deng2009imagenet} to capture the classifiable properties of samples. This method has the drawbacks of neglecting the visual quality of the generated results and prefers models who generate objects rather than realistic scene images, but it can still provide essential diversity information of synthesized samples in evaluating generative models.
\begin{table}[h!]
    \begin{center}
    \resizebox{\columnwidth}{!}{
      \begin{tabular}{c|c|c} 
        \toprule
        \textbf{Model Type} & \textbf{Name} & \textbf{Inception Score}\\
        \hline
        & \textbf{Real Images}& 11.24$\pm$0.11\\
        \hline
        \multirow{3}{*}{Implicit Models}
        &DCGAN & \textbf{6.16$\pm$0.07}\\
        &Improved GAN & 4.36$\pm$0.05\\
        &ALI & 5.34$\pm$0.05\\
        \hline
        \multirow{5}{*}{Descriptive Models}
        &WINN-5CNNs  & 5.58$\pm$0.05\\
        &FRAME~(wl) & 4.95$\pm$0.05\\
        &FRAME & 4.28$\pm$0.05\\
        &wFRAME~(ours,wl) & \textbf{6.05$\pm$0.13}\\
        &wFRAME~(ours) & 5.52$\pm$0.13\\
        \bottomrule
      \end{tabular}
    } 
    \caption{Inception score on datasets CIFAR-10 where 'wl' means training with labels. The IS result of ALI is reported in~\cite{warde2016improving}. IS of DCGAN is reported in~\cite{wang2016learning}, and the result of Improved GAN(wl) is reported in~\cite{salimans2016improved}. WINN's is reported in \cite{Lee_2018_CVPR}. In the Descriptive Model plate, wFRAME outperforms the most methods.
    \label{tab:cifar}} 
    \end{center}
\end{table}
\subsection{Comparison with GANs}
\begin{figure}[h!]
    \centering
    \centerline{\includegraphics[width=\linewidth]{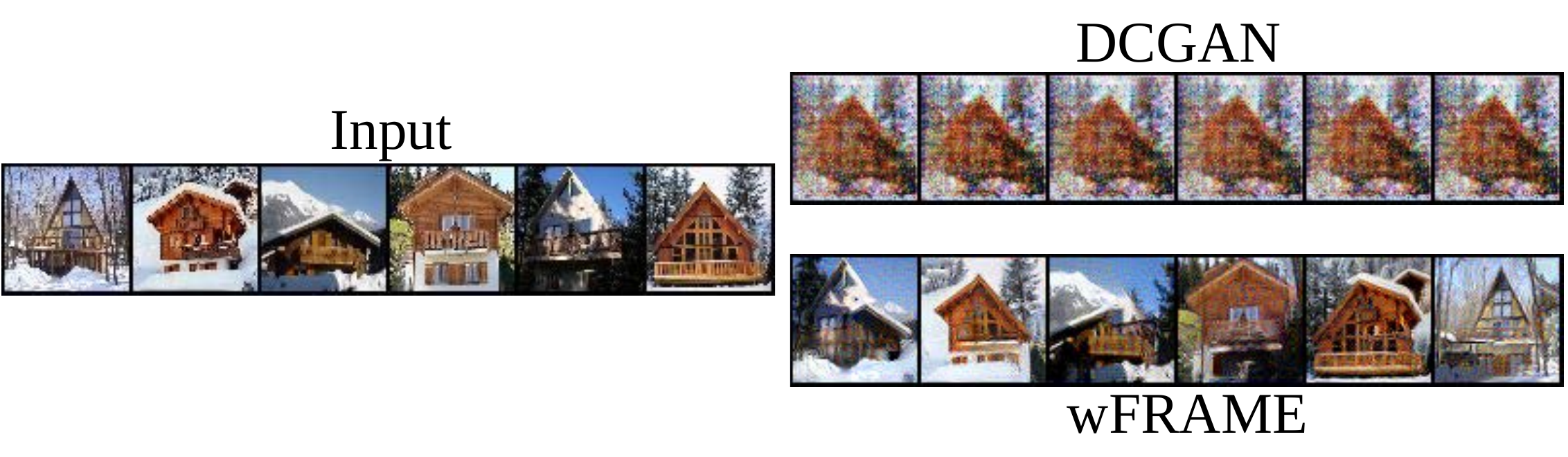}}
    \caption{The first left row is the selected input images from the SUN dataset, the right first row is the random outputs of DCGAN, the right last row is the outputs of our method.\label{fig:worse_gan}}
\end{figure}

We compare FRAME and wFRAME with GAN models implemented on CIFAR-10 via the Inception score in Table~\ref{tab:cifar}. Most GAN-family models achieve pretty high on this score, however, our method is a descriptive model instead of an implicit model. GANs with high scores perform badly in descriptive situations, for example, the image reconstruction task or training on a small amount of data. FRAME can handle most of these situations properly. The performance of DCGAN in modeling mere few images is presented in Fig.~\ref{fig:worse_gan} where for equal comparison, we duplicate the input images several times to the total amount of 10000 to adopt the training environment of DCGAN. The compared wFRAME is trained in our own method. The DCGAN's training procedure is ceased as it converges but still remains collapsed results. 

\subsection{Comparison of FRAME and wFRAME}
From two aspects, we analyze FRAME and wFRAME as a summary of the whole experiments conducted above. As expected, our algorithm is more suitable for synthesizing complex and varied scene images and the resulting images are apparently more authentic compared with FRAME.

\subsubsection{Quality of Generation Improvement.}According to our performances on response distance $R$, the quality of the image synthesis is improved. This measurement is corresponding with the iteration learning process of both FRAME and wFRAME. The learning curves presented in Fig.~\ref{fig:curve} are the observations of the overall datasets synthesis. From the curves can we draw the conclusion that wFRAME converges better than FRAME. The results of generation on CelebA, LSUN-Bedroom and CIFAR-10 in Fig.~\ref{fig:celeba_lsun} and~\ref{fig:cifar_vis} shows that even if the training images are relatively aligned with conspicuous structural information, or with only simple categorical context information, the images produced by FRAME are still abundant with motley noise and twisted texture, while ours are more reasonably mixed, more sensible structured and bright-colored with less distortion.

\subsubsection{Training Steadiness Improvement.}Compared with FRAME as shown in Fig.~\ref{fig:iters} which illustrates the typical evolution of generated samples, we found an improvement on the training steadiness. The generated images are almost identical at the beginning, however, images produced by our algorithm are able to be back on track after 30 iterations while FRAME's deteriorate. Quantitatively in Fig.~\ref{fig:curve}, the curves are calculated by averaging across the whole dataset. wFRAME reaches lower cost on response distance, namely the direct $L_{1}$ critic of filter banks between synthesized samples and target samples is smaller and decreases more steadily. To be more specific, our algorithm has mostly solved the model collapse problem of FRAME for it not only ensures the closeness between the generated samples and ``ground-truth'' samples but also stabilizes the learning phase of the model parameter $\boldsymbol{\theta}$. The three plots clearly show the quantitative measures are well correlated with qualitative visualizations of generated samples. In the absence of collapsing, we attain comparable or even better results over FRAME.

\section{Conclusion}
In this paper, we re-derivatively track the origin of FRAME from the viewpoint of particle evolution and have discovered the potential factors that may lead to the deterioration of sample generation and the instability of model training, i.e, the inherent vanishing problem existing in the minimization of KL divergence. Based on this discovery, we propose wFRAME by reformulating the KL discrete flow in the FRAME to the JKO scheme, and prove through empirical examination that it can overcome the above-mentioned deficiencies. The experiments are carried out to demonstrate the superiority of the proposed wFRAME model and comparable results have shown that it can greatly ameliorate the vanishing issue of FRAME and can produce more visually promising results.
\bibliography{wframe}

\begin{thebibliography}{}

\bibitem[\protect\citeauthoryear{Adams \bgroup et al\mbox.\egroup
  }{2011}]{adams2011large}
Adams, S.; Dirr, N.; Peletier, M.~A.; and Zimmer, J.
\newblock 2011.
\newblock From a large-deviations principle to the wasserstein gradient flow: a
  new micro-macro passage.
\newblock {\em Communications in Mathematical Physics} 307(3):791--815.

\bibitem[\protect\citeauthoryear{Ambrosio, Gigli, and
  Savar{\'e}}{2008}]{ambrosio2008gradient}
Ambrosio, L.; Gigli, N.; and Savar{\'e}, G.
\newblock 2008.
\newblock {\em Gradient flows: in metric spaces and in the space of probability
  measures}.
\newblock Springer Science \& Business Media.

\bibitem[\protect\citeauthoryear{Arjovsky, Chintala, and
  Bottou}{2017}]{arjovsky2017wasserstein}
Arjovsky, M.; Chintala, S.; and Bottou, L.
\newblock 2017.
\newblock Wasserstein generative adversarial networks.
\newblock In {\em International Conference on Machine Learning},  214--223.

\bibitem[\protect\citeauthoryear{Benamou and
  Brenier}{2000}]{benamou2000computational}
Benamou, J.-D., and Brenier, Y.
\newblock 2000.
\newblock A computational fluid mechanics solution to the monge-kantorovich
  mass transfer problem.
\newblock {\em Numerische Mathematik} 84(3):375--393.

\bibitem[\protect\citeauthoryear{Dai, Lu, and Wu}{2014}]{dai2014generative}
Dai, J.; Lu, Y.; and Wu, Y.-N.
\newblock 2014.
\newblock Generative modeling of convolutional neural networks.
\newblock {\em arXiv preprint arXiv:1412.6296}.

\bibitem[\protect\citeauthoryear{Deng \bgroup et al\mbox.\egroup
  }{2009}]{deng2009imagenet}
Deng, J.; Dong, W.; Socher, R.; Li, L.-J.; Li, K.; and Fei-Fei, L.
\newblock 2009.
\newblock Imagenet: A large-scale hierarchical image database.
\newblock In {\em Computer Vision and Pattern Recognition (CVPR), 2009. IEEE
  Conference on},  248--255.
\newblock IEEE.

\bibitem[\protect\citeauthoryear{Duong, Laschos, and
  Renger}{2013}]{duong2013wasserstein}
Duong, M.~H.; Laschos, V.; and Renger, M.
\newblock 2013.
\newblock Wasserstein gradient flows from large deviations of many-particle
  limits.
\newblock {\em ESAIM: Control, Optimisation and Calculus of Variations}
  19(4):1166--1188.

\bibitem[\protect\citeauthoryear{Erbar \bgroup et al\mbox.\egroup
  }{2015}]{erbar2015large}
Erbar, Matthias~anErbar, M.; Maas, J.; Renger, M.; et~al.
\newblock 2015.
\newblock From large deviations to wasserstein gradient flows in multiple
  dimensions.
\newblock {\em Electronic Communications in Probability} 20.

\bibitem[\protect\citeauthoryear{Goodfellow \bgroup et al\mbox.\egroup
  }{2014}]{goodfellow2014}
Goodfellow, I.; Pouget-Abadie, J.; Mirza, M.; Xu, B.; Warde-Farley, D.; Ozair,
  S.; Courville, A.; and Bengio, Y.
\newblock 2014.
\newblock Generative adversarial nets.
\newblock In {\em Advances in Neural Information Processing Systems},
  2672--2680.

\bibitem[\protect\citeauthoryear{Jordan, Kinderlehrer, and
  Otto}{1998}]{jordan1998variational}
Jordan, R.; Kinderlehrer, D.; and Otto, F.
\newblock 1998.
\newblock The variational formulation of the fokker--planck equation.
\newblock {\em SIAM journal on mathematical analysis} 29(1):1--17.

\bibitem[\protect\citeauthoryear{Koller and
  Friedman}{2009}]{koller2009probabilistic}
Koller, D., and Friedman, N.
\newblock 2009.
\newblock {\em Probabilistic graphical models: principles and techniques}.
\newblock MIT press.

\bibitem[\protect\citeauthoryear{Krizhevsky and
  Hinton}{2009}]{krizhevsky2009learning}
Krizhevsky, A., and Hinton, G.
\newblock 2009.
\newblock Learning multiple layers of features from tiny images.
\newblock Technical report, Citeseer.

\bibitem[\protect\citeauthoryear{Krizhevsky, Sutskever, and
  Hinton}{2012}]{krizhevsky2012imagenet}
Krizhevsky, A.; Sutskever, I.; and Hinton, G.~E.
\newblock 2012.
\newblock Imagenet classification with deep convolutional neural networks.
\newblock In {\em Advances in neural information processing systems},
  1097--1105.

\bibitem[\protect\citeauthoryear{Landau and Lifshitz}{2013}]{landau2013course}
Landau, L.~D., and Lifshitz, E.~M.
\newblock 2013.
\newblock {\em Course of theoretical physics}.
\newblock Elsevier.

\bibitem[\protect\citeauthoryear{Lee \bgroup et al\mbox.\egroup
  }{2018}]{Lee_2018_CVPR}
Lee, K.; Xu, W.; Fan, F.; and Tu, Z.
\newblock 2018.
\newblock Wasserstein introspective neural networks.
\newblock In {\em The IEEE Conference on Computer Vision and Pattern
  Recognition (CVPR)}.

\bibitem[\protect\citeauthoryear{Liu \bgroup et al\mbox.\egroup
  }{2015}]{liu2015deep}
Liu, Z.; Luo, P.; Wang, X.; and Tang, X.
\newblock 2015.
\newblock Deep learning face attributes in the wild.
\newblock In {\em Proceedings of the IEEE International Conference on Computer
  Vision},  3730--3738.

\bibitem[\protect\citeauthoryear{Lu, Zhu, and Wu}{2015}]{lu2015learning}
Lu, Y.; Zhu, S.-C.; and Wu, Y.~N.
\newblock 2015.
\newblock Learning frame models using cnn filters.
\newblock {\em arXiv preprint arXiv:1509.08379}.

\bibitem[\protect\citeauthoryear{Montavon, M{\"u}ller, and
  Cuturi}{2016}]{montavon2016wasserstein}
Montavon, G.; M{\"u}ller, K.-R.; and Cuturi, M.
\newblock 2016.
\newblock Wasserstein training of restricted boltzmann machines.
\newblock In {\em Advances in Neural Information Processing Systems},
  3718--3726.

\bibitem[\protect\citeauthoryear{Rubner, Tomasi, and
  Guibas}{2000}]{rubner2000earth}
Rubner, Y.; Tomasi, C.; and Guibas, L.~J.
\newblock 2000.
\newblock The earth mover's distance as a metric for image retrieval.
\newblock {\em International journal of computer vision} 40(2):99--121.

\bibitem[\protect\citeauthoryear{Salimans \bgroup et al\mbox.\egroup
  }{2016}]{salimans2016improved}
Salimans, T.; Goodfellow, I.; Zaremba, W.; Cheung, V.; Radford, A.; and Chen,
  X.
\newblock 2016.
\newblock Improved techniques for training gans.
\newblock In {\em Advances in Neural Information Processing Systems},
  2234--2242.

\bibitem[\protect\citeauthoryear{Simonyan and
  Zisserman}{2014}]{simonyan2014very}
Simonyan, K., and Zisserman, A.
\newblock 2014.
\newblock Very deep convolutional networks for large-scale image recognition.
\newblock {\em arXiv preprint arXiv:1409.1556}.

\bibitem[\protect\citeauthoryear{Szegedy \bgroup et al\mbox.\egroup
  }{2016}]{szegedy2016rethinking}
Szegedy, C.; Vanhoucke, V.; Ioffe, S.; Shlens, J.; and Wojna, Z.
\newblock 2016.
\newblock Rethinking the inception architecture for computer vision.
\newblock In {\em Proceedings of the IEEE conference on computer vision and
  pattern recognition},  2818--2826.

\bibitem[\protect\citeauthoryear{Villani}{2003}]{villani2003topics}
Villani, C.
\newblock 2003.
\newblock {\em Topics in optimal transportation}.
\newblock Number~58. American Mathematical Soc.

\bibitem[\protect\citeauthoryear{Wang and Liu}{2016}]{wang2016learning}
Wang, D., and Liu, Q.
\newblock 2016.
\newblock Learning to draw samples: With application to amortized mle for
  generative adversarial learning.
\newblock {\em arXiv preprint arXiv:1611.01722}.

\bibitem[\protect\citeauthoryear{Warde-Farley and
  Bengio}{2016}]{warde2016improving}
Warde-Farley, D., and Bengio, Y.
\newblock 2016.
\newblock Improving generative adversarial networks with denoising feature
  matching.

\bibitem[\protect\citeauthoryear{Xiao \bgroup et al\mbox.\egroup
  }{2010}]{xiao2010sun}
Xiao, J.; Hays, J.; Ehinger, K.~A.; Oliva, A.; and Torralba, A.
\newblock 2010.
\newblock Sun database: Large-scale scene recognition from abbey to zoo.
\newblock In {\em Computer vision and pattern recognition (CVPR), 2010 IEEE
  conference on},  3485--3492.
\newblock IEEE.

\bibitem[\protect\citeauthoryear{Xie \bgroup et al\mbox.\egroup
  }{2016a}]{xie2016cooperative}
Xie, J.; Lu, Y.; Zhu, S.-C.; and Wu, Y.~N.
\newblock 2016a.
\newblock Cooperative training of descriptor and generator networks.
\newblock {\em arXiv preprint arXiv:1609.09408}.

\bibitem[\protect\citeauthoryear{Xie \bgroup et al\mbox.\egroup
  }{2016b}]{xie2016theory}
Xie, J.; Lu, Y.; Zhu, S.-C.; and Wu, Y.
\newblock 2016b.
\newblock A theory of generative convnet.
\newblock In {\em International Conference on Machine Learning},  2635--2644.

\bibitem[\protect\citeauthoryear{Xie \bgroup et al\mbox.\egroup
  }{2018}]{xie2018learning}
Xie, J.; Zheng, Z.; Gao, R.; Wang, W.; Zhu, S.-C.; and Wu, Y.~N.
\newblock 2018.
\newblock Learning descriptor networks for 3d shape synthesis and analysis.
\newblock {\em arXiv preprint arXiv:1804.00586}.

\bibitem[\protect\citeauthoryear{Xie, Zhu, and Wu}{2017}]{xie2017synthesizing}
Xie, J.; Zhu, S.-C.; and Wu, Y.~N.
\newblock 2017.
\newblock Synthesizing dynamic patterns by spatial-temporal generative convnet.
\newblock In {\em Proceedings of the IEEE Conference on Computer Vision and
  Pattern Recognition},  7093--7101.

\bibitem[\protect\citeauthoryear{Younes}{1989}]{younes1989parametric}
Younes, L.
\newblock 1989.
\newblock Parametric inference for imperfectly observed gibbsian fields.
\newblock {\em Probability Theory and Related Fields} 82(4):625--645.

\bibitem[\protect\citeauthoryear{Yu \bgroup et al\mbox.\egroup
  }{2015}]{yu15lsun}
Yu, F.; Zhang, Y.; Song, S.; Seff, A.; and Xiao, J.
\newblock 2015.
\newblock Lsun: Construction of a large-scale image dataset using deep learning
  with humans in the loop.
\newblock {\em arXiv preprint arXiv:1506.03365}.

\bibitem[\protect\citeauthoryear{Zhu, Wu, and Mumford}{1997}]{zhu1997minimax}
Zhu, S.~C.; Wu, Y.~N.; and Mumford, D.
\newblock 1997.
\newblock Minimax entropy principle and its application to texture modeling.
\newblock {\em Neural Computation} 9(8):1627--1660.

\end{thebibliography}
\bibliographystyle{aaai}

\clearpage
\appendix
\section{A Proofs}
\subsection{Proof of Lemma 1}
\label{lemma1}
With another perspective that under the Gaussian reference measure, FRAME established on morden ConvNet has the piecewise Gaussian property and it's summarized in Proposition~\ref{prop:piecegauss}.
\begin{proposition}
    \label{prop:piecegauss}
    (Reformulation of Theorem 1.) Equation 3 is piecewise Gaussian, on each piece the probability density can be written as:
    \begin{equation}
        \label{eq:piecegauss}
        P(\boldsymbol{x};\boldsymbol{\theta}) \propto \exp[-\frac{1}{4} \Vert \boldsymbol{x}-\boldsymbol{y} \Vert^{2}]
    \end{equation}
\end{proposition}
where $\boldsymbol{y}=\mathbf{B}_{\boldsymbol{\theta}, \delta}=\sum_{k=1}^{K}\delta_{k}\boldsymbol{\theta}_{k}$ is an approximated reconstruction of $\boldsymbol{x}$ in one piece of data space by a linear transformation involving inner-products with model parameter and piecewise linear activation function(ReLu). This proposition implies that different pieces can be regarded as different generating samples acting as Brownian particles.

By Proposition~\ref{prop:piecegauss}, each particle~(image piece) in FRAME has the transition kernel in Gaussian form~(equation~\ref{eq:piecegauss}). It describes the probability of a particle moving from $\boldsymbol{x} \in \mathbb{R}^{d}$ to $\boldsymbol{y} \in \mathbb{R}^{d}$ in time $\tau>0$. Let a fixed measure $\rho_{0}$ such as Gaussian be the initial measure of $n$ Brownian particles $\{\boldsymbol{x}_{0}^{i}\}_{i=1}^{n}$ at time $0$. Sanov theorem shows that empirical measure $\rho_{\tau}=\frac{1}{n}\sum_{i=1}^{n}\delta_{\boldsymbol{x}^{i}_{\tau}}$ of such transition particles satisfy LDP with rate functional $\mathcal{K}(\rho_{\tau}\mid\rho_{0})$, i.e.,
\begin{equation}
    \label{eq:sanov}
    \frac{1}{n}\sum_{i=1}^{n}\delta_{\boldsymbol{x}^{i}_{\tau}} \sim \exp[-\mathcal{K}(\rho_{\tau}\mid\rho_{0})].
\end{equation}

Specially, each Brownian particle has $\vert\mathcal{X}\vert$ internal sub-particles which are independent in different cliques. Let $C$ denote the number of cliques, Cramer's theorem tells us that for i.i.d. RVs $\boldsymbol{x}^{i}(j)$ with common generating function $\mathbb{E}[e^{\Phi(\boldsymbol{\theta})}]$, the empirical mean $\frac{1}{C}\sum_{j=1}^{C}\boldsymbol{x}^{i}(j)$ satisfies LDP with rate functional in the Legendre transformation of $\mathbb{E}[e^{\Phi(\boldsymbol{\theta})}]$,
\begin{equation}
    \label{eq:cramer}
    \frac{1}{C}\sum_{j=1}^{C}\boldsymbol{x}^{i}(j) \sim \exp\Big[\boldsymbol{\theta}-\mathbb{E}[\Phi(\boldsymbol{\theta})]\Big].
\end{equation}
Since the empirical measure of $\boldsymbol{x}_{\tau}^{i}$ is simply the empirical mean of the Dirac measure, i.e., $\delta_{\boldsymbol{x}^{i}_{\tau}}\frac{1}{C}\sum_{j=1}^{C}\boldsymbol{x}_{\tau}^{i}(j)$, then the empirical measure over all particles achieves to
\begin{align*}
    &\frac{1}{n}\sum_{i=1}^{n}\delta_{\boldsymbol{x}^{i}_{\tau}}
    \frac{1}{C}\sum_{j=1}^{d}\boldsymbol{x}^{i}_{\tau}(j)
    \stackrel{Eq.~\ref{eq:sanov}~\ref{eq:cramer}}{\sim}\\
    &  \exp[-\mathcal{K}(\rho_{\tau}\mid\rho_{0})]\cdot\exp\Big[\theta-\mathbb{E}[\Phi(\boldsymbol{\theta})]\Big] \\
    &\propto \exp\Big[\mathcal{K}(\rho_{\tau}\mid\rho_{0})+\mathbb{E}[\Phi(\boldsymbol{\theta})]\Big].
\end{align*}
where the exponent is exactly the KL discrete flow $\mathcal{I}$. Thus the empirical measure of the activation patterns of all those particles satisfies LDP with rate functional $\mathcal{I}_{t}$ in discrete time. \qed

\subsection{Proof of Theorem 1}
\label{thm1}
\textbf{Necessity} can be constructed using MEP via calculating $\partial_{\rho}\mathcal{I}_{t}^{l}$ iteratively:
\begin{equation*}
    \begin{aligned}
        \operatorname*{lim}_{t\to\infty}\rho_{t+1}
            & = \frac{1}{Z} e^{\omega_{t}\cdot\Phi}\rho_{t} \\
            & = \frac{1}{Z} e^{(\omega_{t}+\omega_{t-1})\cdot\Phi}\rho_{t-1} \\ & = \cdot\cdot\cdot=\frac{1}{Z} e^{\Phi(x;\boldsymbol{\theta
                })}q.
    \end{aligned}
\end{equation*}

\textbf{Sufficiency:} Recall the Markov property in Eq.~\ref{prop:piecegauss}, we can write the inner product as sum of feature responses w.r.t. different clique, then shared pattern activation $h=\max(0,\boldsymbol{x}_{i}^{\tau})$ can be approximated by the Dirac measure as
\begin{align*}
    \sum_{k=1}^{K}h(\langle \boldsymbol{x}_{i}^{\tau},\boldsymbol{w}_{k}\rangle+b_{k})
        & \approx \sum_{k=1}^{K} \delta_{\boldsymbol{x}_{i}^{\tau}}\sum_{j=1}^{C}\boldsymbol{x}_{i}^{\tau}(j)\boldsymbol{w}_{k}(j)\\
        &\propto \delta_{\boldsymbol{x}_{i}^{\tau}}\sum_{j=1}^{C}\boldsymbol{x}_{i}^{\tau}(j).
\end{align*}
The result coincides with the empirical measure of $\boldsymbol{x}_{i}^{\tau}$, so the proof of \textbf{sufficiency} turns into the proof of Lemma~\ref{lemma:1} and it was done in Lemma~\ref{lemma:1}. \qed

\section{B More Visual Results}
\begin{figure}[bh]
    \centering
    \centerline{\includegraphics[width=0.68\linewidth]{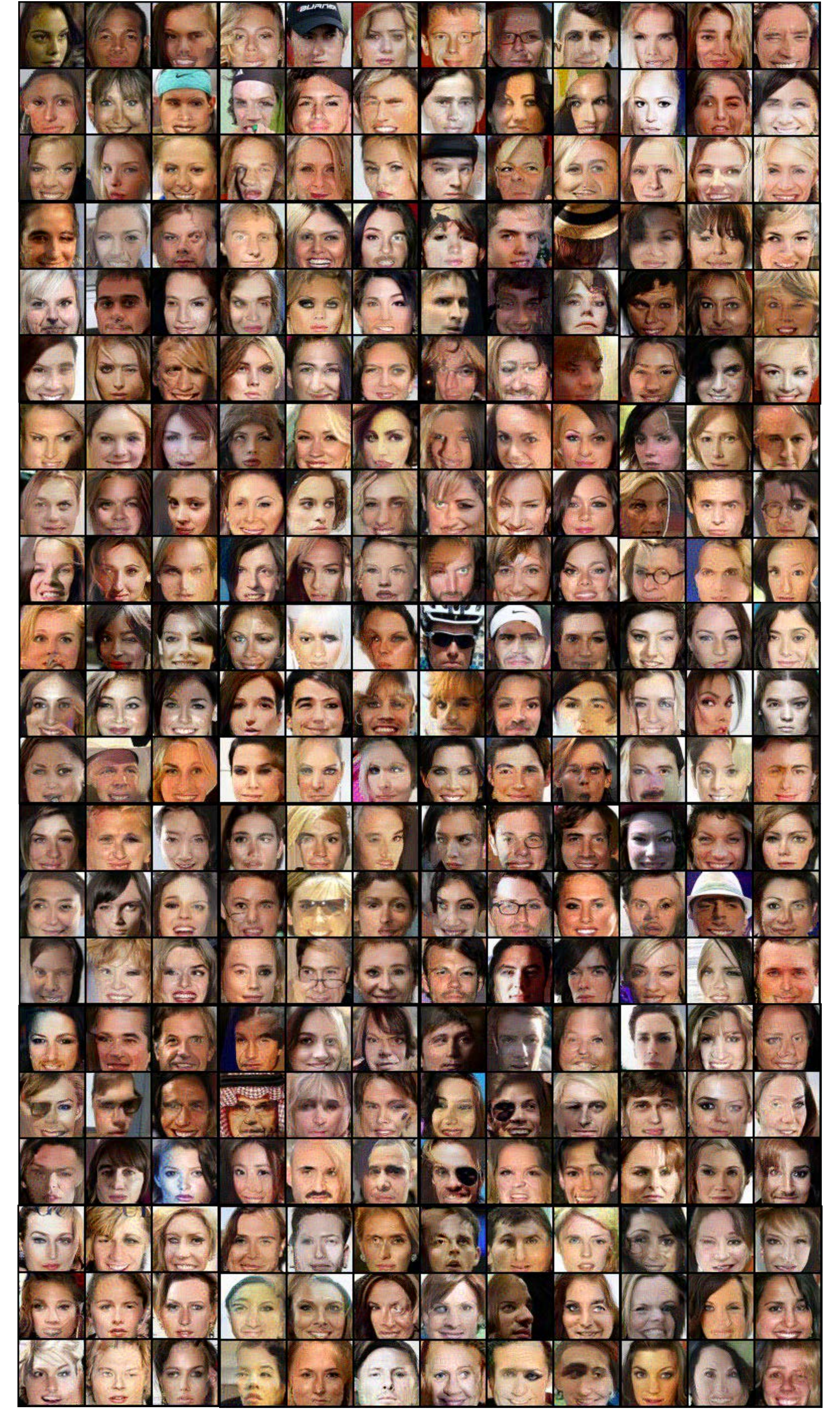}}
    \caption{Synthesized images on celebA.\label{fig:supl_celeba}}
    \end{figure}
    \begin{figure*}[h]
        \centering
        \centerline{\includegraphics[width=0.8\paperwidth]{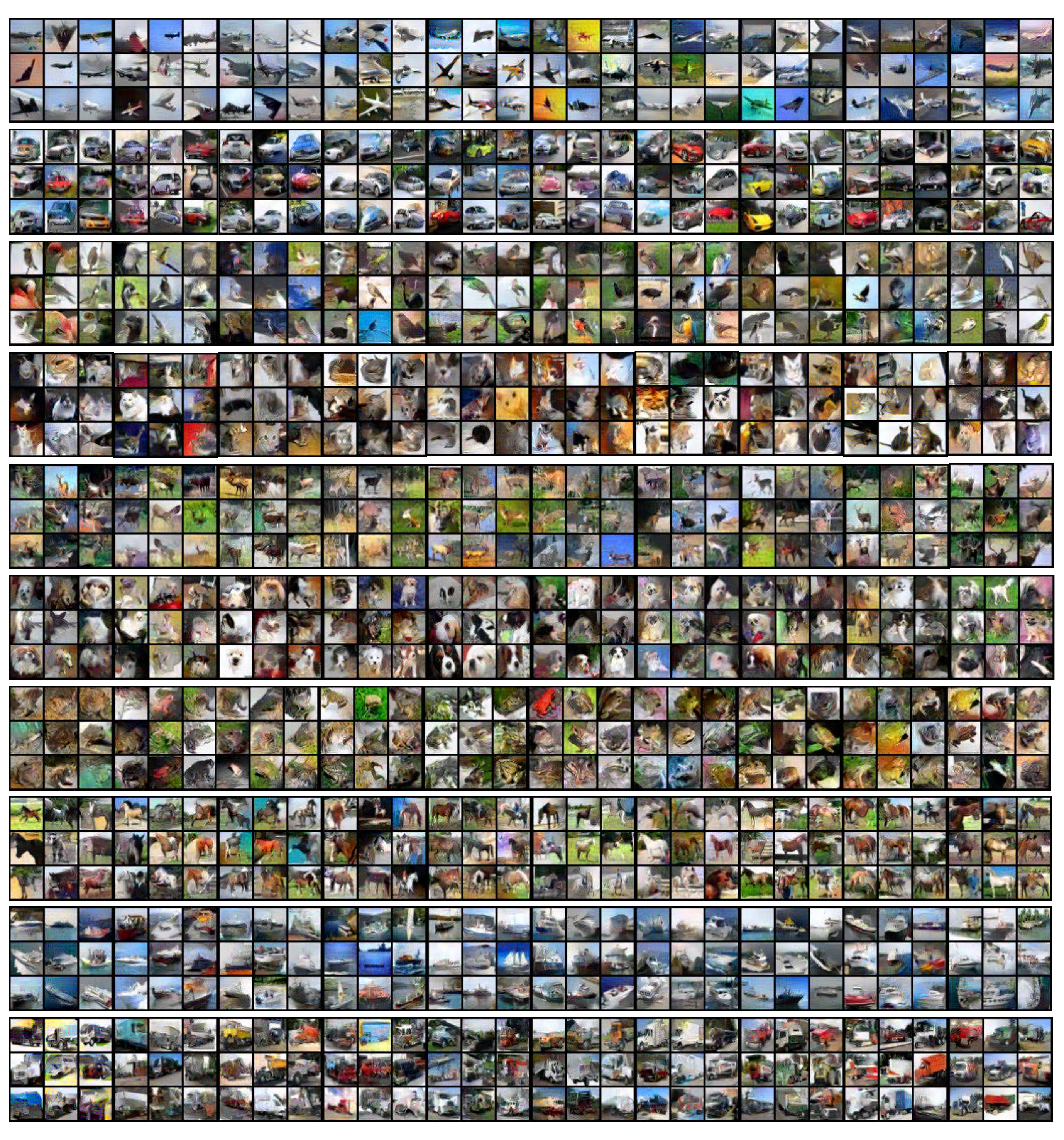}}
        \caption{Synthesized images conditioned on labels in CIFAR-10, each row is of one class \label{fig:supl_cifar10}}
        \end{figure*}
\begin{figure*}[h!]
    \centering
    \centerline{\includegraphics[width=0.8\paperwidth]{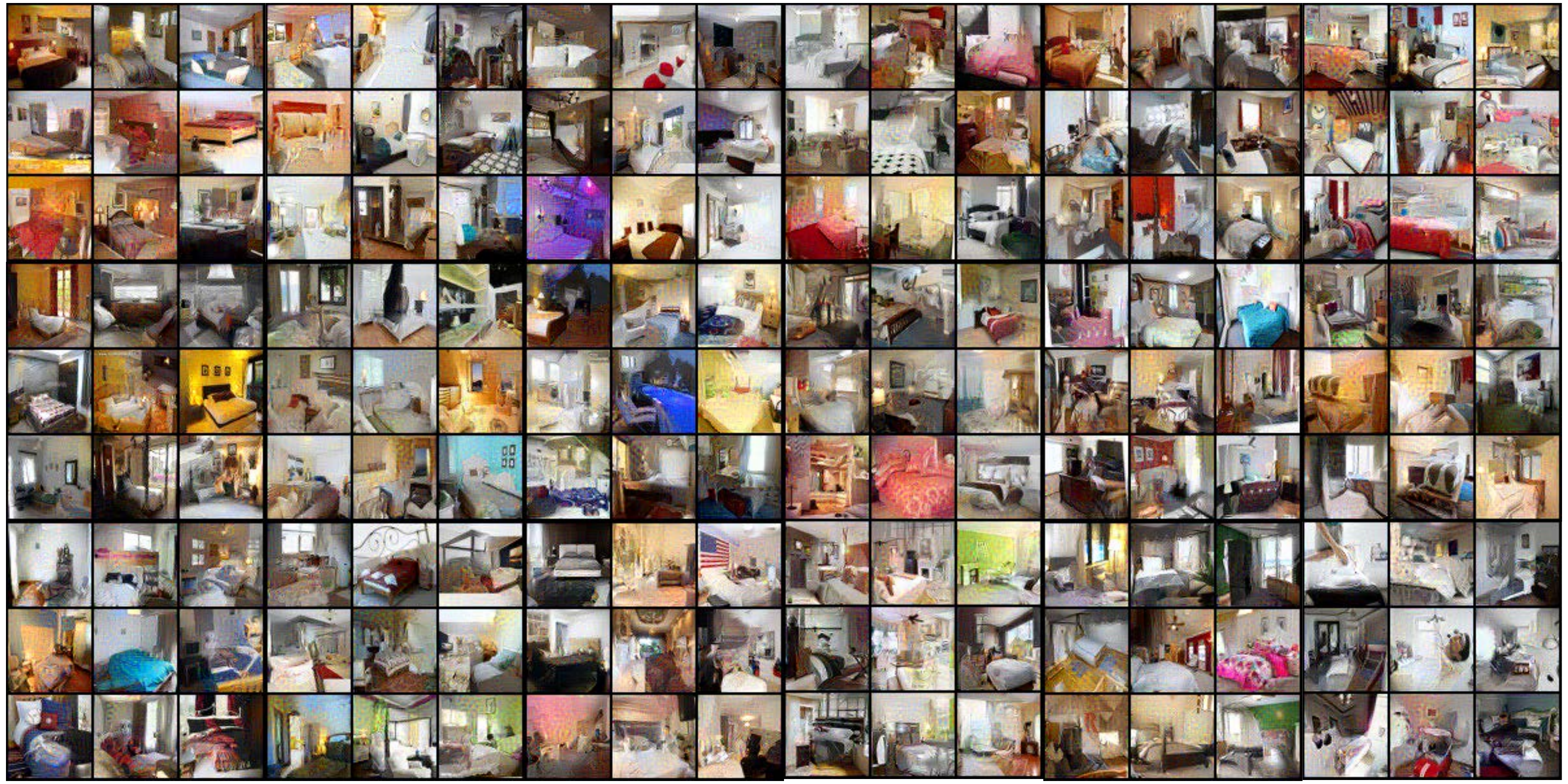}}
    \caption{Synthesized images on LSUN-Bedroom.\label{fig:supl_lsun}}
    \end{figure*}

\end{document}